\documentclass{article}
\usepackage{amsmath}
\usepackage{amssymb}
\usepackage{graphicx} 
\usepackage{multicol}
\usepackage{multirow}

\title{A Multi-view Discourse Framework for Integrating Semantic and Syntactic Features in Dialog Agents}
\author{Akanksha Mehndiratta, Krishna Asawa}

\begin{document}

\maketitle
\begin{abstract}
    Multiturn dialogue models aim to generate human-like responses by leveraging conversational context, consisting of utterances from previous exchanges. Existing methods often neglect the interactions between these utterances or treat all of them as equally significant. This paper introduces a discourse-aware framework for response selection in retrieval-based dialogue systems. The proposed model first encodes each utterance and response with contextual, positional, and syntactic features using Multi-view Canonical Correlation Analysis (MCCA). It then learns discourse tokens that capture relationships between an utterance and its surrounding turns in a shared subspace via Canonical Correlation Analysis (CCA). This two-step approach effectively integrates semantic and syntactic features to build discourse-level understanding. Experiments on the Ubuntu Dialogue Corpus demonstrate that our model achieves significant improvements in automatic evaluation metrics, highlighting its effectiveness in response selection.
    \end{abstract}
\section{Introduction}
 Dialogue systems can be categorized into task-oriented dialogue systems \cite{santhanam2019survey, wu2020tod, quan2020modeling}, and non-task-oriented dialogue systems \cite{xing2018hierarchical, qiu2020structured, lin2020world}, also known as open-domain dialogue systems. Task-oriented dialogue systems aim to assist users in accomplishing specific tasks, such as booking a flight or offering customer support. They rely on predefined dialogs, often enhanced with natural language understanding (NLU) models, to interpret user input and generate responses. As virtual assistants, they excel at handling structured tasks and delivering goal-directed outcomes but lack the flexibility to manage open-ended or non-task-related queries. In contrast, open-domain systems are capable of handling conversations on a wide range of topics that are not restricted to specific tasks or domains. These systems have experienced significant growth due to the availability of large dialogue datasets\cite{moghe2018towards,bi2019fine,li2017dailydialog,zhou2020kdconv} and advancements in deep learning and pre-trained models \cite{radford2018improving, Devlin2019BERTPO}. However, they face challenges in maintaining coherence and accuracy during longer conversations.

In open-domain dialogue systems, retrieval-based agents have gained substantial attention. In multi-turn response selection, a dialogue context is provided as input, and retrieval-based methods aim to develop systems that can identify relevant responses or information from a large database based on the input context. The dialogue context includes both system and user interactions, providing critical details about the topic and focus of the ongoing conversation.
Recent developments in multi-turn retrieval-based agents have focused on language modeling techniques such as sequential models like RNNs and LSTMs \cite{lowe2015ubuntu}, attention-based models \cite{yan2016learning}, and hierarchical architectures (e.g., RNN-RNN \cite{wu-etal-2017-sequential} or CNN-RNN \cite{zhou2016multi}). More recently, transformers and pre-trained models have significantly boosted the performance of dialogue agents. Although pretrained models have enhanced response quality, they are limited by high computational costs and difficulties in managing long-term context.

This paper introduces a novel approach to integrate the dialogue context with discourse-level knowledge by focusing on the syntactic and semantic significance of the input context in a conversation for response selection. The proposed method exploits the multi-view nature of textual data in two key ways and utilizes Canonical Correlation Analysis (CCA)\cite{hotelling1992relations} to develop discourse-level knowledge.
The model begins by developing utterance level representations for each utterance in the input context and for candidate responses using Multi-View Canonical Correlation Analysis (MVCCA) to capture a multi-modal perspective of an utterance. Next, it interprets the utterance level representations of an utterance and its surrounding dialogue turns as two sets of variables that have a joint multivariate normal distribution and exploits Canonical Correlation Analysis (CCA) to learn shared intent representations as correlation coefficients. Unique intent representations are treated as discourse tokens. The degree of similarity between the discourse tokens for the dialogue context and the candidate responses determines how aligned a response is.

By identifying semantic similarity using discourse tokens, the proposed Multi-view Discourse Framework(MVDF) aims to highlight the importance of filtering out suitable context in fetching responses that are contextually relevant and coherent, ensuring smooth dialogue progression. Experiments conducted with the Ubuntu dataset demonstrate that the proposed model improves automatic evaluation metrics, proving its effectiveness.

\section{Related Work}
A generalized framework for retrieval-based dialogue agents utilizes information retrieval (IR) techniques to rank a set of pre-built candidate responses and select the most appropriate one from the top-ranked candidates. Chatbots built on this framework have garnered significant attention within the information retrieval community, due to their ability to generate fluent and informative responses\cite{liu2019machine, yan2016learning, yan2017joint}. 

Approaches in this field have evolved from sequence-to-sequence and transformer models to methods using pre-trained models such as BERT \cite{Devlin2019BERTPO}. Early research in retrieval-based agents focused on discourse modeling in multiparty dialog systems. Models trained on annotated datasets developed for spoken language, such as the STAC corpus \cite{Asher2016DiscourseSA} and the Molweni corpus\cite{li2020molweni}, largely focus on detecting links between discourse dependencies and classifying discourse relations. Discourse modeling-based frameworks include approaches such as MST \cite{afantenos2015discourse}, ILP \cite{perret2016integer}, and Deep Sequential \cite{Shi2018ADS}.

The prosperity of the deep learning paradigm led to the conception of multi-turn conversational systems where a dense and continuous vector represents the textual information. The modeling framework can be classified into the following categories: (1) representation-based models \cite{lowe2015ubuntu, yan2016learning, zhou2016multi} and (2) interaction-based models \cite{zhou2018multi, zhang2018modeling, yuan2019multi, wu-etal-2017-sequential, tao2019multi, tao2019one, lu2019constructing}. Representation-based models focus on generating a representation of the encoding for the input document and query and obtaining a score based on the degree of matching between them. The interaction-based models first build a matching framework between input query and document to compute interaction matrices, then using the interaction matrix as input, a CNN or variant of CNN is utilized to output the similarity score. 

More recently, pre-trained models have seen significant success in natural language processing tasks. These high-capacity contextual language models are front-runners in almost every task such as Question Answering (QA) and Natural Language Inference (NLI) \cite{Devlin2019BERTPO}. Although these pre-trained models excel at capturing semantic relationships and dependencies within text, these systems are computationally intensive, and their pre-training objective hampers their ability to manage long-term context\cite{mehndiratta2021non}. 

\section{Canonical Corelation Analysis}

This study exploits CCA to achieve the following objectives:

\subsection{Multiview Learning based Language Modeling}
CCA has been used vividly to analyze datasets with multiple views to learn shared latent structures between the views, enabling better integration of multimodal data. This study aims to leverage an extension of CCA, called Multiview Canonical Correlation Analysis (MCCA)\cite{rupnik2010multi}, to perform exploratory analysis on multi-view data\cite{zhao2017multi}. Multi-view data refers to datasets where multiple sets of variables (views) describe the same set of entities. MCCA extends standard CCA by enabling the analysis of more than two sets of variables simultaneously, allowing for the exploration of correlations across multiple views or perspectives of the same data. By incorporating several views, MCCA can capture richer relationships within the data.

To obtain utterance level representations that effectively integrate semantic and syntactic features, this study utilizes the different views of an utterance to better capture the structure, coherence, and flow of a conversation. In dialogue systems, it is crucial to understand the meaning behind utterances, and MCCA facilitates this by correlating multiple representations (or views) of an utterance to develop an integrated encoding in a common subspace. The utterance level representations in this study are based on the following views of each utterance in the input context.
\begin{enumerate}
    \item Contextual Representations: Capture the meaning of words in an utterance.
    \item Positional Representations: Capture the position of each word within an utterance.
    \item Syntactic Representations: Capture the syntactic role of each word in an utterance, using part-of-speech information.
\end{enumerate}

To develop utterance level representations, the defined views are analyzed using MCCA. The goal is to introduce a method for learning language models that estimates a hidden state representation for words based on their position, context, and part of speech. Specifically, we compute the dominant canonical correlations across multiple modalities of a word to estimate the associated latent state. By applying CCA to estimate a latent space for each word based on its semantic and syntactic attributes, a vector is computed that represents each word type. This is done by using the right singular values of MCCA to map from the multiple word spaces to the latent state space of size k.

\subsection{Feature Extraction Using Hidden State Interpretation}
\begin{figure}[ht]
\centering
    \includegraphics[width=0.6\linewidth]{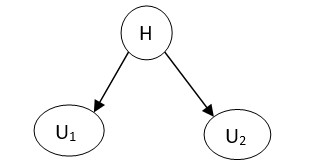}
    \caption{Hidden state interpretation of CCA on two random variables}
    \label{fig:latent}
\end{figure}

CCA has the capability to extract features that are maximally correlated between two datasets, which is useful for exploration of semantically relevant relationships between two sets of variables[\cite{bach2005probabilistic,mehndiratta2024discovering}] represented by the graphical model figure \ref{fig:latent}. Discovering discourse-level knowledge involves exploring different utterances of a dialogue to better capture its structure, coherence, and flow. In dialogues, understanding context is essential for interpreting the meaning behind a conversation. CCA can assist by correlating utterances with their surrounding utterances to generate discourse encodings or vectors that represent the relationships across these utterances\cite{mehndiratta2025modeling}. These discourse encodings are maximally correlated, ensuring that the most relevant parts of the conversation are processed. This approach reduces computational overhead while preserving the critical elements of the dialogue.

\section{Multi-view Discourse Framework(MVDF)}
\subsection{Model Definition} In a multi-turn dialog agent, the input is a context C and the goal of the model is to score a response R by utilizing the discourse knowledge that is relevant to the context C. Let the input context be represented as C = {U\textsubscript{1}, U\textsubscript{2}, . . . , U\textsubscript{t}}; where t is the number of input context utterances of a conversation; Let R be a candidate response.

\subsection{Model Architecture}

\subsubsection{Define The Views} The input utterance is represented as M-view data X = {X\textsubscript{1}, X\textsubscript{2} $ .... $ X\textsubscript{m}} where each utterance has m views, each defined in its own vector space. We have n instances in X representing n word tokens in the utterance. Hence each view is represented in a space $\in$ $\mathbb{R}$\textsuperscript{n x v\textsubscript{i}}, where v\textsubscript{i} is the vector space of the i\textsuperscript{th} view. The study defines multiple representations of an utterance to gain a multi-modal perspective, that includes
\begin{enumerate}
\item Contextual representation: To understand the meaning of each utterance based on context the study develops contextual embedding using pre-trained models.
\item Positional representation: To capture the flow of conversation based on the sequence of words in an utterance the study develops positional encodings for each word.
\item Syntactic representation: To incorporate the structure of a utterance, representations for part-of-speech (POS) tagging for each word are utilized.
\end{enumerate}

\begin{figure}[ht]
    \centering
    \includegraphics[width=\linewidth]{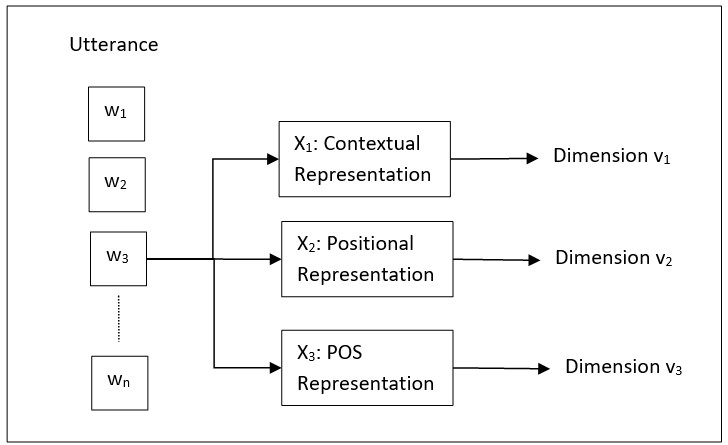}
    \caption{Word token representation in a multi-view space}
    \label{fig:mvr}
\end{figure}

\subsubsection{Multi-view Learning Based Utternace Level Representations} 
\begin{figure}[ht]
    \centering
    \includegraphics[width=\linewidth]{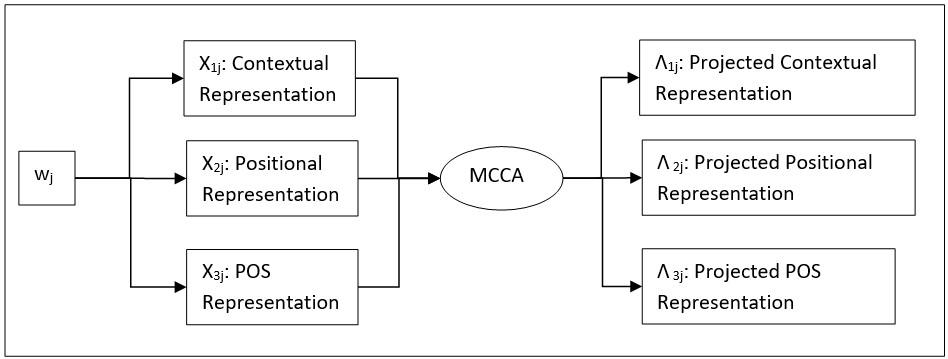}
    \caption{Overview of the Multi-view learning based language modeling}
    \label{fig:mvlearning}
\end{figure}

Let us assume that an utterance consists of n tokens {w\textsubscript{1}, w\textsubscript{2}, $...$, w\textsubscript{n}}. Let X\textsubscript{1}, X\textsubscript{2} and X\textsubscript{3} be a set of vectors representing a token w\textsubscript{j}. Here, X\textsubscript{1} consists of the contextual vector representation of the word w\textsubscript{i}  defined in a space with dimension as v\textsubscript{1}, X\textsubscript{2} consists of the vector representation of the position of w\textsubscript{j} defined in a space with dimensionality as v\textsubscript{2} and X\textsubscript{3} consists of the vector representation of POS of w\textsubscript{j} defined in a space with dimensionality as v\textsubscript{3}, as illustrated in figure\ref{fig:mvr}. For each random vector X\textsubscript{i} with dimension v\textsubscript{i}, MCCA constructs a univariate random variable taking a linear combination of its components: $\beta$\textsubscript{i}X\textsubscript{i}. Then it computes the pairwise correlation coefficients for each pair of variables, in order to identify the vectors $\beta$\textsubscript{i} that maximize the sum of all pairwise correlations. Therefore, MCCA constructs projection lambda, as shown in figure \ref{fig:mvlearning}, that projects (X\textsubscript{1}, X\textsubscript{2} and X\textsubscript{3}) to the most correlated lower-dimensional subspace of size k = min(v\textsubscript{1}, v\textsubscript{2} and v\textsubscript{3}) using the following equation
\begin{equation}
\Lambda\textsubscript{1}, \Lambda\textsubscript{2}, \Lambda\textsubscript{3} = MCCA(X\textsubscript{1}, X\textsubscript{2}, X\textsubscript{3})
\end{equation}
Foster et al.\cite{foster2008multi} stated that the best linear predictor derived from each individual view is approximately as effective as the best linear predictor using all views together. Hence the feature encoding for a word token are derived from $\Lambda$\textsubscript{1}. Utterance Level Representations are designed using the feature encodings of all word token from an utterance.

\subsubsection{Feature Extraction Based Discourse Framework}
To effectively develop long term dependencies in a conversation and elevate its role this study presents a discourse learning framework using spectral methods. The foundational units of discourse called Discourse tokens are determined throughout the course of a conversation. Unique discourse tokens are collected from a conversation to develop a discourse level knowledge. To perform a discourse level understanding it is important to understand how the contextual information introduced at the start of the conversation relates to later turns. 

With the utterance level representations developed for the first utterance and the utterance that follows, CCA analyzes both to output projections, called intents H, as defined in Equation below.

\begin{equation}
\delta\textsubscript{U1}, \delta\textsubscript{u2} = CCA(U\textsubscript{1}, U\textsubscript{2})
\end{equation}

\begin{figure}[ht]
    \includegraphics[width=\linewidth]{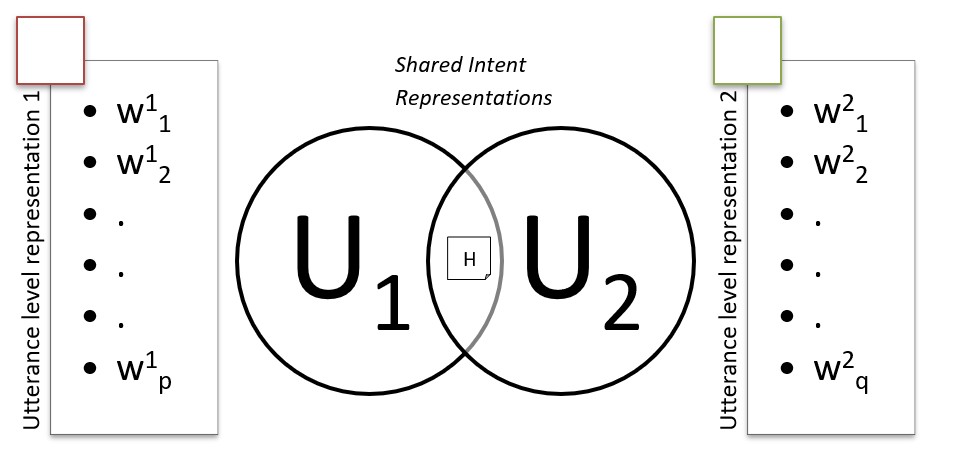}
    \caption{Hidden state interpretation of CCA on two utterances to learn shared intent}
    \label{fig:MVDF1}
\end{figure}

Let U\textsubscript{1} and U\textsubscript{2} be matrices with k columns in which the j\textsuperscript{th} row represents the feature encoding learned for the j\textsuperscript{th} word token in an utterance. U\textsubscript{1} contains p word representations, while U\textsubscript{2} represents the matrix for the second turn, containing q word representations. The canonical variates, $\delta$\textsubscript{1} and $\delta$\textsubscript{2}, each contain word representations that provide a redundant estimate of the same hidden state H, as shown in figure \ref{fig:MVDF1}. The first a projections learned through CCA are referred as intents and unique intents are collected as discourse tokens. These discourse tokens contain essential information regarding the underlying associations and hidden structures from the utterance pair, referencing crucial contextual information for the ongoing conversation.
Further intents and discourse tokens are learned by performing CCA between the discourse developed and the next utterance. Long-term dependencies are developed by analyzing the relationship between discourse level information introduced at the beginning of a conversation with the later utterances. 

\subsubsection{Response Selection}
Utilizing MCCA, each candidate response is modeled to generate an utterance-level representation. A matching score is then computed by measuring the cosine similarity between the utterance level representation of a response and the discourse tokens identified after processing the entire context. Based on the matching score the candidate responses are ranked and the model selects the most appropriate one from the top-ranked candidates.

\subsubsection{Evaluation metrics}
To perform a quantitative analysis, the study evaluates the performance of the model using Recall@k. Given the candidate responses for each context in the test set, R\textsubscript{10}@k indicates whether the positive response is present among the top-k retrieved responses from the 10 candidates. To evaluate the quality of the responses selected, this study performs a qualitative analysis using popular qualitative evaluation metrics in Natural Language Processing (NLP) namely perplexity, BLUE score, Rouge Score and lexical diversity.

\section{Experiment}
\subsection{Dataset} The Ubuntu Dialog Corpus is a large-scale dataset that was specifically designed for training and evaluating models in the context of dialogue systems. It is often used in research related to retrieval-based dialogue systems, where the goal is to find the most relevant response from a set of possible candidate responses based on the user's input. The dataset is based on IRC chat logs from Ubuntu-related channels, where users interact to solve technical issues or seek help with using Ubuntu. 

The dataset consists of conversational pairs (dialogue turns), typically between a user asking a question and an expert or another user providing a solution or response. The conversation is sampled from a random turn where the prior dialogue history or previous exchanges leading to the random turn is termed as context while the remaining dialogues as the ground truth response. Each context-response pair is annotated for relevance, therefore some random responses are selected and labeled as incorrect.

\subsection{Language modeling using MCCA}
To perform language modeling and generate a representation for each word in an utterance we define various modalities of the text as follows: 
Contextual representation: Contextual embedding for each word token in an utterance is developed using BERT.
Positional representation: Positional token is appended by the end of each word token to generate a positional embedding using BERT.
Syntactic representation: POS embeddings are developed using OneHotEncoder method defined in SKlearn library. 
MCCA is defined in mvlearn, an open-source Python library, as the method mcca = MCCA(n$\_$ components) and mcca.fit(X\textsubscript{1}, X\textsubscript{2}, X\textsubscript{3}) to perform Multiview canonical correlation analysis for any number of views. It also defines the mcca.transform(X\textsubscript{1}, X\textsubscript{2}, X\textsubscript{3}) method that returns one dataset per view of the shape (n$\_$samples, n$\_$ components). Here n$\_$ samples is the number of input samples so in this case number of word tokens in an utterance while n$\_$ components is limited to size k = min(v\textsubscript{1},v\textsubscript{2},v\textsubscript{3}). Utterance level representations are generated by using the projections learned for the contextual embedding view.

\subsection{Discourse framework using CCA}
\begin{figure}[ht]
    \includegraphics[width=\linewidth]{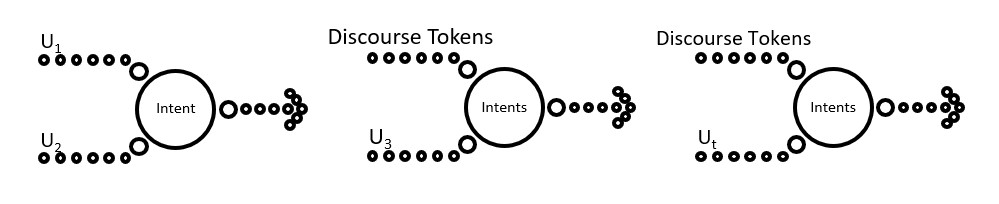}
    \caption{Overview of the proposed discourse framework using CCA}
    \label{fig:MVDF2}
\end{figure}

The Utterance level representations generated for each utterance in the context are utilized to develop intents using CCA as described in figure\ref{fig:MVDF2}. CCA is defined in both SKLearn and MVlearn open-source Python libraries. To perform feature extraction using the CCA method defined in the library are CCA(n$\_$ components) and cca.fit(U\textsubscript{1}\textsuperscript{T}, U\textsubscript{2}\textsuperscript{T}) to fit the model to input data and to output the projections (latent variable pairs) $\Delta$\textsubscript{U1}, $\Delta$ \textsubscript{U2} of shape (n$\_$ samples, n$\_$ components) per utterance level representation method cca.transform(X,Y) transforms the input utterance level representation by maximizing the correlation between them. Intents are developed by using the projections learned from both the utterance level representations. Here n$\_$ samples is equal to the number of features that are used to represent a word tokens in an utterance i.e. k, while the number of components is limited to size l = min(number of word tokens in U\textsubscript{1}, number of word tokens in U\textsubscript{2} )

\subsection{Baselines}
Several models have been developed to enhance multi-turn response selection. The proposed model is an extension of \cite{mehndiratta2025modeling} where the model proposed performs feature extraction for collecting intents and discourse tokens using CCA. This work extends the proposed model by enhancing the utterance level representations to capture syntactic characteristics and integrating them in creating a discourse level understanding.

\subsection{Result and Analysis}
Table \ref{tab:tab1} presents the performance of the proposed MVDF model, evaluated across three benchmark evaluation metrics. The results indicate that our model is highly competitive compared to the baseline models. 

To qualitatively analyze the framework, the Rouge score, BLEU score and perplexity are evaluated for the top-k-ranked responses. The results are presented in Table \ref{tab:tab123}. These metrics assess the richness of the selected responses. A lower perplexity score suggests a better understanding of language, while a BLEU or Rouge score of 1 indicates a perfect match between the selected response and the target response.

These findings highlight the effectiveness of integrating multiple representations through a multi-representation fusion network, which better captures the characteristics of dialogue. The improvement demonstrates the importance of filtering the relevant discourse tokens from context before matching. By learning a limited yet relevant set of discourse tokens, MVDF significantly enhances semantic relevance and coherence between internal utterances, ultimately improving response selection performance.

\begin{table}
\caption{Results of the models on the experiment using Recall\textsubscript{n}@k}
    \label{tab:tab1}
    \centering
    \begin{tabular}{|c|c|c|c|c|} 
       \hline
 Model & Recall\textsubscript{all}@20 & Recall\textsubscript{all}@10 & Recall\textsubscript{all}@5 & Recall\textsubscript{all}@3 \\ \hline
 CCA+Global\cite{mehndiratta2025modeling} & 69 & 62 & 57 & 57 \\ \hline
 CCA+Local\cite{mehndiratta2025modeling} & 71 & 64 & 60 & 59 \\  \hline
 MVDF & 73 & 66 & 60 & 60 \\ \hline
    \end{tabular}
    
\end{table}

\begin{table}
\caption{Results of the models on various popular evaluation metrics in NLP}
    \label{tab:tab123}
    \centering
    \begin{tabular}{|p{2.3cm}|p{1.8cm}|p{1.5cm}|p{1.5cm}|p{2cm}|p{2cm}|p{2cm}|} 
       \hline
 Model & Perplexity & BLEU Score & Rouge - 1 Score & Rouge - L Score & Distinct N Unigram & Distinct N Bigram\\ \hline
 CCA+Global\cite{mehndiratta2025modeling} & 23.1891 & 0.3345 & 0.1469 & 0.1365 & 0.1800 & 0.4554\\ \hline
 CCA+Local\cite{mehndiratta2025modeling} & 28.3874 & 0.3192 & 0.1476 & 0.1374 & 0.1774 & 0.4490\\  \hline
 MVDF & \textbf{18.012} & \textbf{0.3921} & \textbf{0.1834} & \textbf{0.1732} & \textbf{0.1853} & \textbf{0.4657} \\ \hline
    \end{tabular}
    
\end{table}

\section{Conclusion}
This paper introduces a Multi-view discourse framework for response selection in open-domain dialogue systems by integrating dialogue context with discourse-level knowledge. Leveraging Multi-View Canonical Correlation Analysis (MVCCA), the proposed method captures both syntactic and semantic relationships within multi-turn conversations to develop utterance level representation. It then learns discourse tokens that represent the relationships between an utterance and its surrounding turns in using Canonical Correlation Analysis (CCA). The model enhances response retrieval by aligning the response with the discourse tokens learned, improving contextual relevance and coherence.

Experimental results on the Ubuntu dataset demonstrate the effectiveness of this approach, showcasing significant improvements in automatic evaluation metrics. The findings highlight the potential of discourse-aware methods in enhancing retrieval-based dialogue systems. Extending this approach to diverse conversational domains and datasets and investigating the impact of various CCA variants specifically Deep CCA, a deep-learning based framework, would be a valuable direction for future research.

\bibliography{name}
\bibliographystyle{abbrv}
\end{document}